# An Agile Method for Implementing Retrieval-Augmented Generation Tools in Industrial SMEs

BOURDIN Mathieu*[a,e], NEUMANN Anas[b,c], PAVIOT Thomas[a,d],

PELLERIN Robert[b,c], LAMOURI Samir[a,e]

*Corresponding author: mathieubourdin@hotmail.fr

[a]LAMIH CNRS/Université Polytechnique Hauts-de-France, Campus Mont Houy, 59313 Valenciennes Cedex 9, France
[b]CIRRELT, 2920 chemin de la Tour, pavillon André Aisenstadt, bureau 3520, 11290 Montréal, Canada
[c]Polytechnique Montréal, 2500 Chemin de Polytechnique Montréal, 11290 Montréal, Canada
[d]Meedia, 15 rue Glais Bizoin, 35000 Rennes, France
[e]Arts et Métiers Institute of Technology, 151 boulevard de l'Hôpital, 75013 Paris, France

Abstract

Retrieval-Augmented Generation (RAG) has emerged as a powerful solution to mitigate the limitations of Large Language Models (LLMs), such as hallucinations and outdated knowledge. However, deploying RAG-based tools in Small and Medium Enterprises (SMEs) remains a challenge due to their limited resources and lack of expertise in natural language processing (NLP). This paper introduces EASI-RAG: Enterprise Application Support for Industrial RAG, a structured, agile method designed to facilitate the deployment of RAG systems in industrial SME contexts. EASI-RAG is based on method engineering principles and comprises well-defined roles, activities and techniques. The method was validated through a real-world case study in an environmental testing laboratory, where a RAG tool was implemented to answer operators' queries using data extracted from operational procedures. The system was deployed in less than a month by a team with no prior RAG experience and was later iteratively improved based on users' feedback. Results demonstrate that EASI-RAG supports fast implementation, high user adoption, delivers accurate answers, and enhances the reliability of underlying data. This work highlights the potential of RAG deployment in industrial SMEs. Future works include the need for generalization across diverse use cases and further integration with fine-tuned models.

**Keywords:** Industry 4.0; Retrieval Augmented Generation (RAG); Natural Language Processing (NLP); Large Language Models (LLM); SME

# 1. Introduction

Since the release of OpenAI's ChatGPT at the end of 2022, public and research interest in language models has surged dramatically. Within just a few years, numerous new language models have emerged, including Meta's LLaMa, Google's Gemini, OpenAI's GPT-4, Anthropic's Claude, Mistral AI's Mixtral, and the DeepSeek LLM family.

These tools are increasingly adopted by companies to optimise tasks that previously required significant manual effort. However, one notable limitation is their tendency to hallucinate [1]. Language models sometimes generate completely incorrect information. This issue poses a significant barrier to the use of LLMs across many business applications that demand reliable information [2]. Furthermore, outputs from LLMs are not directly explainable, as these models operate as black boxes due to their vast architectures containing billions of parameters. It has also been demonstrated that large language models are not well-suited for knowledge storage: Mousavi et al. [3] showed that among 24 LLMs tested, including some recently released, none could correctly answer more than 80% of simple questions related to current events. This is because their "knowledge" reflects the data they were trained on and may therefore be outdated.

To avoid these issues, it is necessary to integrate as much information as possible into the LLM before generating answers. This can be done in two ways: Retrieval-Augmented Generation (RAG) or finetuning. RAG consists in retrieving relevant information from one or several databases linked to the query before generating a response using the retrieved data. This approach both reduces the risk of hallucination and allows for easier updating of system knowledge by maintaining a database rather than retraining an entire LLM [4]. Fine-tuning, on the other hand, consists of adapting a pre-trained language model to a specific domain or task by retraining it on a curated dataset relevant to the target application. This technique enables the model to internalise domain-specific terminology and patterns, improving performance on specialised queries.

Bourdin et al. [5] highlighted the lack of scientific literature offering guidance for deploying NLP solutions in industrial contexts. In particular, no existing guide helps industrial SMEs, which usually lack NLP expertise and generally have limited resources [6][7], navigate the wide range of available options for implementing previously mentioned solutions.

Fine-tuning a language model requires specific domain-expertise as well as significant computational resources, which makes this option hardly compatible with SMEs, where both human and computational resources are usually more limited than in large companies [6]. Moreover, using a fine-tuned LLM requires retraining every time the underlying data is updated, which is a constraint for SMEs, whose agility is a competitive advantage compared to larger enterprises. For these reasons, RAG-based solutions are much better suited for use in SMEs than fine-tuning approaches.

The objective of this work is to bridge the identified gap in scientific literature by developing a method to implement RAG-based solutions within the context of industrial SMEs. To this end, we constructed the EASI-RAG method: Enterprise Application Support for Industrial RAG, an agile step-by-step method for deploying RAG tools in industrial SMEs. The methodology proposed in this work was developed using an inductive approach based on an extensive review of the scientific literature. By synthesizing common patterns and principles from prior studies on Retrieval-Augmented Generation (RAG) systems, we derived a structured, step-by-step framework tailored for industrial applications. This review, initiated in a previous paper [5] and extended in the present work, forms the basis for developing the proposed method. EASI-RAG considers the company's objectives as well as the constraints and characteristics of its data to select the most appropriate RAG solutions for a given industrial application.

This work addresses Natural Language Processing (NLP) from two perspectives: both user queries and database documents are in natural language. It focuses on industrial use cases involving information synthesis or retrieval from text, such as maintenance, diagnostics, customer support, regulatory compliance, summarization, and employee training. The scope is limited to textual data, excluding temporal data (e.g., time series forecasting) and numerical analysis. Although image-based data is not the primary focus, the method is compatible with tools such as OCR. RAG is not required in small-context scenarios, as some LLMs now process over 100,000 tokens without preprocessing. However, Liu et al. [8] showed that too much context can impair retrieval, especially for mid-context information, even at 2,000 tokens. RAG becomes valuable when the knowledge base exceeds several thousand tokens. The target use cases thus involve contexts where the volume of textual data justifies the use of retrieval mechanisms.

The model was subsequently tested and validated through a concrete RAG application case. The model was used to implement a system that uses RAG to answer operators' questions based on the company's operating procedures. The system aims to save operators time when searching through dense documentation and improve accuracy by facilitating access to the correct information.

The remainder of the paper is structured as follows: Section 2 reviews existing work in this field. Section 3 presents the EASI-RAG model. Section 4 then describes the application case developed with our industrial partner, followed by a result discussion in Section 5. Finally, Section 6 summarises the main conclusions of this study.

## 2. Related work

Several studies explore the use of Retrieval-Augmented Generation (RAG) in industrial contexts. Chen et al. [9] introduce an interactive knowledge management (IIKM) system to assist technicians with technical repairs and internal policy inquiries. Chaudhary et al. [10] designed a Llama-based chatbot for Continuous Integration and Continuous Delivery (CI/CD)-related questions in a telecom company, leveraging RAG for document-specific accuracy. Wan et al. [11] developed a hybrid system combining Knowledge Graphs and vector retrieval for smart manufacturing Q&A. Löwhagen et al. [12] developed a chatbot leveraging Retrieval-Augmented Generation (RAG) to provide real-time responses to technicians' queries in the context of problem-solving during train commissioning. Du et al. [13] proposed LLM-MANUF, a multi-model framework that merges outputs from fine-tuned LLMs to enhance decision-making in manufacturing. Ren et al. [14] presented RACI, a RAG-based model for extracting causes from aviation accident reports, demonstrating high performance on noisy real-world data. Ma et al. [15] employ RAG in combination with knowledge graphs to develop a reasoning tool aimed at supporting complex fault diagnosis. Lastly, Yang et al. [16] detailed the deployment of a RAG-based virtual assistant (RAGVA) at Transurban, offering practical engineering insights and outlining eight key development challenges. These examples highlight RAG's growing role in improving information access, decision support, and automation across industrial sectors.

More general papers offer broader analyses of RAG implementation. Cheng et al. [17] provide a comprehensive review of RAG, examining its core components—retrieval mechanisms, generation processes, and their integration. They propose a taxonomy that classifies RAG approaches, from basic retrieval-augmented models to advanced systems incorporating multimodal inputs and reasoning capabilities. Similarly, Gao et al. [18] present an organised overview of the three key elements of RAG: retriever, generator, and augmentation techniques, detailing the main technologies associated with each. Li et al. [19] examine the relationships between generative AI usage and the performance of supply chains in Chinese companies. They demonstrate a positive correlation between the use of generative AI and supply chain performance, and provide practical guidance on implementing generative AI across supply chains. Wamba et al. [20] adopt a similar approach with English and American companies. They investigate the benefits associated with Gen-AI in Operations and Supply Chain Management and demonstrate that the integration of Gen-AI leads to the enhancement of the overall supply chain performance, particularly when coupled with organizational learning. Finally, Wang et al. [21] explore existing RAG methods and their potential combinations to identify optimal practices. Based on extensive experimentation, they propose deployment strategies that effectively balance performance and efficiency.

Finally, some articles address the implementation of emerging technologies in Small and Medium-sized Enterprises (SMEs). Kong et al. [22] propose a Federated Learning framework to overcome the issues of a lack of data to train Supply Chain Financing models (SCF) in SMEs. Hansen et al. [23] analysed 30 SMEs in their digitalization process and developed a theoretical model of the determinants leading to smart manufacturing, with a focus on the competencies required within the firm to successfully achieve digital transformation. Zheng et al. [24] analyse the different stages of digital transformation process in SMEs through interviews. Their work reveals different phases of data-driven digital transformation in industrial SMEs and indicates preconditions for state transitions in the context of manufacturing SMEs. Lastly, Battistoni et al. [25] analyse how digital technologies support the development of information processing capabilities in Italian SMEs. They propose an Industry 4.0 layer-based classification, providing managers and policymakers with insights on enabling effective digital transformation.

However, although the previously cited papers cover parts of the topic we aim to address, none of them present a method for implementing a RAG-based tool within industrial SMEs. A **method** is a "goal-oriented systematic approach, which helps to resolve theoretical and practical tasks" [26]. As stated by Zellner, a method must include five mandatory elements: Activities, Techniques, Roles, Results, and Information Model [27]. The cited papers present either detailed solutions on specific application cases or provide comprehensive descriptions of existing RAG tools, but without specifying how to choose among the various available options. Our contribution distinguishes itself from cited works by presenting a generic and reusable method applicable to a wide range of RAG-based problem settings beyond a single use case, in the context of industrial SMEs.

Table 1 summarises the aspects addressed by previous works. None of the cited papers covers the entire targeted scope. In particular, it is noteworthy that none of the papers specify the roles associated with each step. Existing models focus on the technical description of the implemented solutions but do not indicate the people who should be involved at each stage.

| Paper | Title | Mandatory Elements of a Method | | | | | Context | | |
|---|---|---|---|---|---|---|---|---|---|
| | | Activities | Techniques | Roles | Results | Info. model | RAG | Industry | SMES |
| [9] | Application of retrieval-augmented generation for interactive industrial knowledge management via a large language model | ◐ | ○ | ○ | ● | ○ | ● | ● | ○ |
| [10] | Developing a Llama-Based Chatbot for CI/CD Question Answering: A Case Study at Ericsson | ● | ● | ○ | ● | ○ | ● | ● | ○ |
| [11] | Empowering LLMs by hybrid retrieval-augmented generation for domain-centric Q&A in smart manufacturing | ● | ● | ○ | ● | ○ | ● | ● | ○ |
| [12] | Can a troubleshooting AI assistant improve task performance in industrial contexts? | ● | ● | ○ | ● | ○ | ● | ● | ○ |
| [13] | LLM-MANUF: An integrated framework of Fine-Tuning large language models for intelligent Decision-Making in manufacturing | ● | ● | ○ | ● | ◐ | ○ | ● | ○ |
| [14] | Retrieval-Augmented Generation-aided causal identification of aviation accidents: A large language model methodology | ● | ● | ○ | ● | ● | ● | ◐ | ○ |
| [15] | A knowledge-graph enhanced large language model-based fault diagnostic reasoning and maintenance decision support pipeline towards industry 5.0 | ● | ● | ○ | ● | ○ | ● | ● | ○ |
| [16] | RAGVA: Engineering retrieval augmented generation-based virtual assistants in practice. | ● | ◐ | ○ | ● | ● | ● | ◐ | ○ |
| [17] | A Survey on Knowledge-Oriented Retrieval-Augmented Generation | ● | ● | ○ | ● | ○ | ● | ○ | ○ |
| [18] | Retrieval-Augmented Generation for Large Language Models: A Survey | ● | ● | ○ | ● | ○ | ● | ○ | ○ |
| [19] | Generative AI-enabled supply chain management: The critical role of coordination and dynamism | ◐ | ◐ | ◐ | ○ | ○ | ○ | ● | ● |
| [20] | Are both generative AI and ChatGPT game changers for 21st-Century operations and supply chain excellence? | ◐ | ◐ | ◐ | ○ | ○ | ○ | ● | ● |
| [21] | Searching for Best Practices in Retrieval-Augmented Generation | ● | ● | ○ | ● | ● | ● | ○ | ○ |
| [22] | A federated machine learning approach for order-level risk prediction in Supply Chain Financing | ● | ● | ○ | ● | ○ | ○ | ● | ● |
| [23] | Technology isn't enough for Industry 4.0: on SMEs and hindrances to digital transformation | ○ | ◐ | ◐ | ◐ | ○ | ○ | ● | ● |
| [24] | A dual evolutionary perspective on the Co-evolution of data-driven digital transformation and value proposition in manufacturing SMEs | ○ | ◐ | ◐ | ● | ○ | ○ | ● | ● |
| [25] | Adoption paths of digital transformation in manufacturing SME | ○ | ◐ | ◐ | ◐ | ○ | ○ | ● | ● |

*Legend:*

● = The paper fully addresses this aspect

◐ = The paper partially addresses this aspect

○ = The paper does not address this aspect at all

*Table 1: Structured evaluation of related works*

# 3. EASI-RAG method

This section presents the newly proposed EASI-RAG approach: a method for implementing RAG-based tools in Industrial SMEs. Following Denner et al.'s definition of a method [28], if offers a systematic procedure, including guidelines and step-by-step strategies, to implement solution. The proposed method is divided into five blocks (see Figure 1):

- **Initial design** (block 1): Selection of technical solutions for each stage of the RAG process;
- **Evaluation** (block 2): Define the performance indicators and the targets, and perform initial evaluation;
- **Error analysis and correction** (block 3): analyse deviations from defined targets and correct them;
- **Integration** (block 4): Integrate the software into the production process; and
- **Feedback gathering** (block 5): Gather user feedback and continuously improve the software.

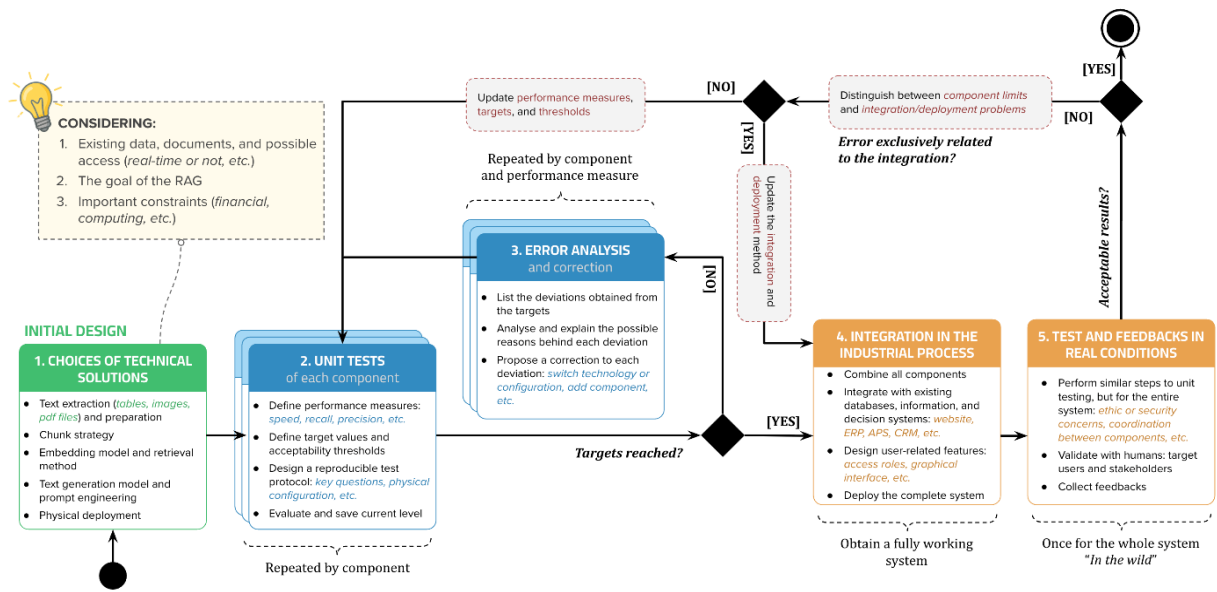

Figure 1 : General representation of the proposed method in five blocks: initial setup, individual evaluation, error correction, integration, test in real conditions, and feedback loops

In the following sections, the five blocks are broken down into activities. For each activity, techniques, roles, and results are specified. Finally, a synthesis of inputs and outputs is provided in the information model (Figure 2).

The method proposed in this paper is based on Agile development principles [29]. The adopted strategy is to quickly deliver an initial, functional RAG-based solution, then refine it iteratively through two iteration loops (Figure 1). This strategy enables rapid tool deployment and strong adaptability to the use case, which are key to SMEs [30]. It also allows to focus on limiting components, saving time and resources, which is especially critical in the resource-constrained context of SMEs [6] [7]. Finally, the method emphasises individuals and their interactions by indicating which roles are required for each activity. The method defines four roles:

- **User:** End-user of the tool; if numerous, a representative sample should be involved;
- **Data expert**: individual with extensive knowledge of the data. Typically, people responsible for updating the data and ensuring its validity;
- **Process Owner** refers to the individual responsible for the process affected by the implementation of the tool, usually the manager of the relevant department;
- **Developer** refers to a member of the IT team responsible for the development of the tool.

This role-centric design addresses a gap in computer science research, which often focuses mainly on technical aspects while overlooking user involvement (see Table 1, "Related Works").

## 3.1 Initial Design of the RAG Process

RAG process is typically broken down into several stages: data retrieval and chunking, vectorisation, chunk retrieval, prompt engineering, and response generation.

### 3.1.1 Data retrieval and chunking

The first activity of our method is the data retrieval and chunking. This step should be performed by data experts and developers. This activity consists in converting files into a usable format, cleaning the data, and segmenting them into chunks. While the first two can be carried out without any specific RAG knowledge, chunking can be more complex due to the large number of available options. The most common strategies are fixed-size chunking [4], element-based chunking [31], sentence-based or paragraph-based chunking [21], sentence window chunking [32], recursive chunking [33], hierarchical chunking [34], and semantic chunking [35]. Comprehensive analysis comparing different chunking methods exist in the literature [32] [34] [36], but as an initial approach, we propose to segment the data simply based on their size.

For short documents (e.g., maintenance reports, incident reports, emails), chunking is generally unnecessary; a single document can be treated as one chunk. For longer documents, chunking should primarily follow the semantic structure of the document if possible (sections and subsections; hierarchical chunking). If not feasible, recursive chunking allows to split documents by paragraphs and sentences until reaching a given token limit. In this case, small chunks (64-128 tokens) are preferable for tasks involving short, fact-based answers whereas larger chunks (512–1024 tokens) are better suited for tasks requiring descriptive or technical responses [36]. These methods allow to preserve semantic coherence and avoiding splitting ideas across chunks.

### 3.1.2 Vectorisation

The vectorisation step should be performed by developers. Two broad families of vectorisers are commonly used:
- count-based methods, also called "sparse" methods, such as TF-IDF or BM25, which represent documents based on term frequencies in a sparse vector [37]; and
- dense vector approaches, also called "embeddings", such as Sentence-BERT or E5, where documents are embedded into a continuous vector space using neural language models [38].

Sparse vectors enable precise keyword-based matching, particularly beneficial in industrial contexts where technical terminology is often crucial. In contrast, dense vectors offer improved contextual understanding.

For sparse vectorisation, the most common approaches are TF-IDF and BM25. For dense vectorisation, a wide range of options is available, making the selection of the most appropriate vectorisation strategy complex [39]. One resource that can help guide this decision is the Massive Text Embedding Benchmark (MTEB) [40]. This benchmark evaluates text embedding models across diverse datasets and tasks, including retrieval, making a valuable starting point to identify candidates for RAG applications. Once the candidates have been found, it is necessary to test a few options on its specific use case. However, vectoriser performance varies by dataset [39] and chunking strategy [34]. Testing multiple options on the specific use case is therefore important to select the most suitable vectoriser.

### 3.1.3 Retrieval of relevant chunks

Stakeholders for the chunk retrieval definition are data experts and developers. The main topic here is to define the number of chunks to retrieve. Finding the best balance in the number of retrieved chunks is significant: retrieving too few chunks may result in the absence of essential contextual elements needed to answer the query accurately, conversely providing an excessive amount of context implies increased response times, higher computational resource consumption, and can reduce the models ability to find relevant information [34]. Even with "small" contexts of 2,000 tokens, language models can miss some information depending on their location in the prompt [8]. A good starting point to mitigate this risk is therefore to limit the context provided to the language model to a few thousand tokens at first (2,000–3,000 tokens). This amount can be increased later depending on the language model used. However, keeping an initial context size limited allows easier control over retrieved chunks and reduces computational cost.

**3.1.4 Prompt engineering**

Stakeholders involved in the prompt engineering activity are users, data experts, and developers. Prompt engineering plays a critical role in obtaining high-quality outputs from large language models (LLMs).

To achieve the most reliable responses, prompts should be formulated with clear and detailed instructions, ideally providing an explicit example of the expected output within the prompt [41]. White et al. [42] present a prompt pattern catalogue to enhance LLM performance by adding simple instructions—e.g., adopting a persona or rules to follow, to specify for example how to respond when lacking information or facing multiple plausible answers. For tasks that require to combine elements or to reason over the data, breaking down complex tasks into sub-tasks can lead to better performance [43] [44]. An easy technique to do so is to include a phrase to explicitly ask for that in the prompt (for example, add *"Think step by step"* in the prompt) [43].

**3.1.5 Language Model selection**

Only the developers and the process owner are involved at this stage. As some language models can be resource-intensive, the process owner must be involved in the case where investment in better hardware is required.

Data confidentiality is crucial when selecting a language model, particularly in industrial companies. For sensitive content, local deployment is recommended to reduce the risk of data leakage or misuse [45]. SMEs, which typically operate with fixed, limited infrastructure, should prioritise lightweight models with acceptable inference times in this case. A practical approach would be to test first several LLMs in a cloud environment, using non-sensitive anonymised data, before potentially investing. If confidentiality is not an issue, proprietary models via APIs are more cost-effective: high-quality models like GPT-4 Turbo or Claude 3 Haiku cost about $1 per million tokens (August 2025), making them affordable even for SMEs.

Other criteria can be taken into consideration when choosing the language model:
- Context window: The model should support sufficiently long inputs to capture relevant content;
- Training objectives: models finetuned on tasks similar to the use case should be preferred; and
- Language/domain coverage: Models trained on corpora relevant to the industrial context (technical manuals, engineering documentation) and in the same language should be preferred.

Table 2 summarises the information about all the activities in the initial RAG pipeline design.

| Activity | Roles | Step(s) | Techniques | Outputs |
|---|---|---|---|---|
| Data retrieval and chunking | Data experts Developers | Accessing data Content pruning Chunking | Element-based, hierarchical, recursive chunking | Available data in segmented chunks |
| Vectorisation | Developers | Obtaining vector representations of chunks | Sparse vectorisations Dense vectorisations Benchmarks and testing | Vector representation of chunks |
| Retrieval of relevant chunks | Data experts Developers | Retrieving relevant chunks for a given query | Real case testing | RAG model retrieving X relevant chunks |
| Prompt engineering | Users Data experts Developers | Engineering the optimal prompt for desired output | Provide explicit examples Give specific guidelines | Prompt template for the language model |
| Language Model selection | Process owner Developers | Choosing the language model for the RAG pipeline | Choosing based on criteria (confidentiality, window size, training, …) | Complete RAG pipeline (untested) |

*Table 2: Summary of the activities for the initial RAG pipeline design*

## 3.2. Evaluation of the RAG Process

After designing the initial RAG pipeline, it is crucial to define its evaluation, involving all key stakeholders: users, data experts, process owners, and developers. The goal is to identify objectives and how to measure success. Evaluation covers response quality—whether the system answers correctly—and performance aspects like inference time, response style, and cost.

To assess response quality, the first step is to build a test query set. The test query set should cover all targeted question types (e.g., multimodal data, complex chains of thought) and include unanswerable questions to test handling of unknowns. Linking test queries to the document excerpt(s) required to answer with the data experts is not mandatory, but is recommended to automate the pipeline evaluation. While automated generation of test queries is possible [46], manual query creation through user collaboration is recommended because LLM are sensitive to phrasing [1]. Criteria for a good answer must be defined as well: high recall (all expected information) and high precision (no additional information) are usual indicators, but are not the only indicator existing (maximum acceptable inference time, required response style, robustness to prompt variations, maximum response generation cost, etc.). Some criteria can be assessed via automatic metrics (BLEU, ROUGE, BERTScore), usually requiring reference answers to compare with. However, automatic metrics might lack reliability in nuanced situations and lack explainability, they are therefore recommended only if the number of test cases is high and alongside manual evaluations.

After defining all the relevant evaluation metrics, a performance target must be set for each criterion and the RAG tool should be tested. Table 3 summarises the information for the RAG pipeline evaluation.

| Activity | Roles | Step(s) | Techniques | Outputs |
|---|---|---|---|---|
| Performance criteria definition | Users Data experts Process owner Developers | Defining the criteria to evaluate the implemented pipeline on the concrete use case | Manual evaluation criteria Automatic tools (BLEU, ROUGE, BertScore) | Performance criteria |
| RAG pipeline testing | Users Data experts Developers | Perform initial evaluation | Evaluation of previously defined performance indicators | Tested RAG pipeline |

*Table 3: Summary of the activities for the RAG pipeline evaluation*

## 3.3. Error analysis and correction

At this stage, key performance indicators for the project have been identified and measured. The next step is to analyse the gaps between actual performance and targets, and to implement the necessary corrections.

**3.3.1 Error analysis**

First, it is essential to examine the deviations from the established objectives. This analysis should be conducted collaboratively by users, data experts, process owners, and developers.

The initial priority is to assess the quality of the system's responses. Using the predefined test questions, each incorrect answer should be reviewed to pinpoint where the failure occurred in the RAG pipeline. Several questions must be considered to this end: Was the information present in the documents? Was it in a supported format? Did the model respond based on prior knowledge? Was the response style appropriate? And so forth. Table 4 outlines all diagnostic questions to locate potential failures (first three columns). Once all errors are reviewed, a Pareto analysis can highlight the most frequent issues to prioritise corrective actions. Note that a single error may involve multiple combined failures. This step reflects the value of an agile approach: focusing first on the most critical weaknesses helps accelerate progress and avoid wasting time on already-satisfactory components.

Regarding the resolution of encountered issues, all individuals involved in the project may be potentially affected, depending on the stage at which our RAG tool is malfunctioning. The following paragraphs describe techniques that can be employed to circumvent the various challenges that may arise during this analysis.

**3.3.2 Incomplete data / Data access issue**

If the data are incomplete or if the system encounters difficulties accessing certain types of data (e.g., figures or tables), several solutions can be considered. If the data are available as images, a multimodal RAG approach can be employed to extract the relevant information [47]. Additionally, recent studies have explored combining RAG with query generation to retrieve data stored in SQL databases [48].

**3.3.3 Inadequate chunking**

To improve the performance of the chunking component, basic approaches are to test alternate chunking strategies (recursive chunking, element-based, hierarchical chunking) or to add overlap between chunks. More advanced approaches involve dividing documents based on semantic similarity (semantic chunking) with another LLM [49] or to combine several chunking strategies [50] [51].

**3.3.4 Chunk retrieval**

Several techniques can enhance retrieval performance. Adjusting the context size (number and length of chunks) is the main lever, as too much or too little context can harm effectiveness. Testing or combining different types of vectorisers or reranking the top-k retrieved chunks through a relevance scoring method can also improve retrieval quality [52]. Eventually, Hypothetical Document Embeddings (HyDE) can generate synthetic document representations from queries to improve retrieval accuracy [53] and child-parent retrieving can facilitate more context-aware retrieval by considering hierarchical relationships within the data [54].

**3.3.5 Unknown vocabulary**

Handling unknown vocabulary is a common topic in industrial contexts. For vocabulary unknown to an **embedding** model, an effective strategy is to implement hybrid retrieval [11], combining dense and sparse representations of texts: sparse vectors for keyword-based matching and dense vectors for contextual understanding. Adding a vocabulary clarification step before the retrieval process using a domain-specific dictionary can also help [55]. For vocabulary unknown to the **language model**, providing specialised vocabulary lists in the prompt or in a separate document in the data can help.

**3.3.6 Answer based on prior knowledge (hallucination)**

Responses generated based on prior knowledge rather than the provided context—commonly referred to as hallucinations—pose significant challenges in RAG systems. Several techniques exist to mitigate this issue: ask the model to explicitly cite the document excerpts used to generate the answer [42] [56], employ grounded models or models fine-tuned on question-answering, or incorporating a fact-checking step prior to delivering the final response [57].

**3.3.7 Inadequacy relevance to the question**

This issue arises when the model misunderstands user intent. To overcome it, query rewriting can clarify intent before retrieval and generation [58]. Another solution can be to generate multiple candidate answers and evaluating them with a relevance metric, like RAGAS' answer relevance metric—comparing artificial questions from answers to the real query—to select the most accurate response [59].

**3.3.8 Lack of logical coherence**

Several techniques have been proposed to improve logical coherence in answers. Chain-of-Thought (CoT) prompting [60] guides the model to generate intermediate reasoning steps. Least-to-Most prompting [61] breaks down complex problems into smaller, manageable subproblems to solve sequentially in a coherent manner. Plan-and-Solve [62] introduce an explicit planning phase before generation, allowing the model to outline a strategy to tackle the query.

### 3.3.9 Lack of consistency in responses

Lack of consistency in answers is another frequent issue when dealing with LLMs. Solutions to this issue include lowering the LM's temperature toward zero to produce more deterministic outputs [63]. Other approaches include refining the prompt to provide the most precise and most explicit instructions possible to guide the model toward consistent behavior [42] or generating multiple responses for the same query and selecting the most consistent one [64].

### 3.3.10 Improper output format

Commonly recommended approaches to prevent poor output formatting are to specify the model's role and desired response style within the prompt [42]. Additionally, supplying examples of well-formatted responses for specific questions helps guide the model toward producing outputs that meet formatting expectations, as demonstrated in [65].

Table 4 summarises all possible actions to undertake based on the encountered issues (last column).

|  | Diagnostic Question | Issue | Metric for Quantification | Actions to correct |
|---|---|---|---|---|
| Retrieval | Are the data required to answer present in the documents? | Incomplete data | Manual analysis | Complete the data; Add new data sources |
| Retrieval | Are all retrieved chunks relevant? | Chunk retrieval | Context precision* | Reduce the amount of context (quantity or size of retrieved chunks); revise the vectorisation method |
| Retrieval | Are all relevant chunks present? | Chunk retrieval | Context recall* | Increase the amount of context; revise the vectorisation method; use reranking strategies; add child-parent retrieval |
| Retrieval | Is the information accessible to the tool? (table, image) | Data access issue | Manual analysis | Implement multimodal RAG systems; use an agent model capable of executing machine-readable actions |
| Retrieval | Is any excerpt split in two during chunking? | Inadequate chunking | Manual analysis | Add overlap; revise chunking strategy; consider hybrid chunking approaches; implement child-parent retrieval |
| Retrieval | Are there terms unknown to the vectoriser model? | Unknown vocabulary | Manual analysis | Integrate sparse vectorisation techniques; employ a synonym dictionary to expand vocabulary coverage |
| Generation | Did the model respond using prior knowledge? | Answer based on prior knowledge | Faithfulness | Apply grounding techniques within the prompt; incorporate a fact-checking step prior to returning a response to the user |
| Generation | Did the answer address the question? | Inadequate relevance to the question | Answer relevance | Use a LM to reformulate the query; generate multiple answers and pick best one based on answer relevance |
| Generation | Are there terms unknown to the language model? | Unknown vocabulary | Manual analysis | Include relevant vocabulary lists directly in the prompt to enhance the model's understanding |
| Generation | Did the LM lack the logic in the provided elements? | Lack of logical coherence | Manual analysis | Utilise structured prompting techniques (CoT, Least-to-Most prompting, Plan-and-Solve) |
| Generation | Is the quality of the obtained response repeatable? | Lack of consistency in responses | Prompt agreement | Reduce the LM temperature; refine prompt; generate multiple answers and keep the most frequent one |
| Generation | Is the style or format of the response appropriate? | Improper output format | Manual analysis | Refine prompt to specify the desired format; provide examples of expected answers |

*Automatic analysis requires having identified in advance the specific chunks to be retrieved for each question

*Table 4: Diagnostic questions for analysing incorrect responses of the RAG tool*

The error analysis and correction activities are summarised in Table 5.

| Activity | Roles | Steps | Techniques | Outputs |
|---|---|---|---|---|
| Error analysis | All | Analyse the root cause(s) of all non-achieved performance criteria | Many (see Table 4) | Root cause analysis of non-achieved targets |
| Error correction | All | Correcting the causes for not achieving the defined objectives | Many (see Table 4) | Corrected RAG pipeline |

Table 5: Summary of the error analysis and correction activities

## 3.4. Integration into the production process

The stakeholders involved at this stage include users, data experts, process owners, and developers, although not all are required for every aspect. Due to the wide range of possible configurations, providing an exhaustive guide for this step is challenging. However, at a minimum, the following elements must be defined:
- **Location of the RAG tool and access**: in particular, whether certain documents should be restricted to specific user groups;
- **Graphical user interface**: should be designed collaboratively by developers and users to ensure usability and functionality;
- **Document updates within the database**: the frequency of updates should be defined —either at fixed intervals or manually triggered. This step typically involves by the process owner and data expert; and
- **Potential integrations with other tools**.

## 3.5. Feedback gathering and continuous improvement

At this stage, all stakeholders are involved. The goal is to collect and leverage feedback to drive continuous software improvement, following the Agile approach. Iteration loops facilitate testing and allow for rapid incorporation of results into subsequent development cycles, while also enhancing collaboration between developers and designers [66]. Short, frequent feedback cycles guide both functional and ergonomic updates [67]. Integrating real-time feedback into interactive systems also reinforces user confidence, a critical aspect, especially when introducing tools based on emerging technologies [68]. For these exchanges to be successful, communication must be bidirectional between users and developers, and not only from users to developers: it developers providing information to users about implemented changes is equally important [69]. Additionally, assessing the impact of each incorrect answer with users and prioritise based on both issue frequency and business impact is critical in agile project management [69]. In sum, the systematic collection and integration of user feedback—through an iterative and collaborative process—is a key success factor in ensuring the adoption and scalability of a tool deployed in a professional environment.

At this stage, if new needs are identified by users, the process returns to the requirements definition stage, updates the evaluation criteria accordingly, and proceeds again through all subsequent steps (see Figure 1).

Table 6 summarises the integration and feedback gathering activities.

| Activity | Roles | Steps | Techniques | Outputs |
|---|---|---|---|---|
| Integration | Users<br>Data experts<br>Process owner<br>Developers | Integrate the designed RAG pipeline to the production | Depending on the use case | Operational RAG integrated to production |
| Feedback gathering and continuous improvement | Users<br>Process owner | Defining how to collect and process feedback | Periodic collaborative reviews with short interval<br>Prioritization and structuring of issues | Prioritised feedback |

*Table 6: Summary of the integration and feedback gathering activities*

Figure 2 presents the complete information model of the proposed method.

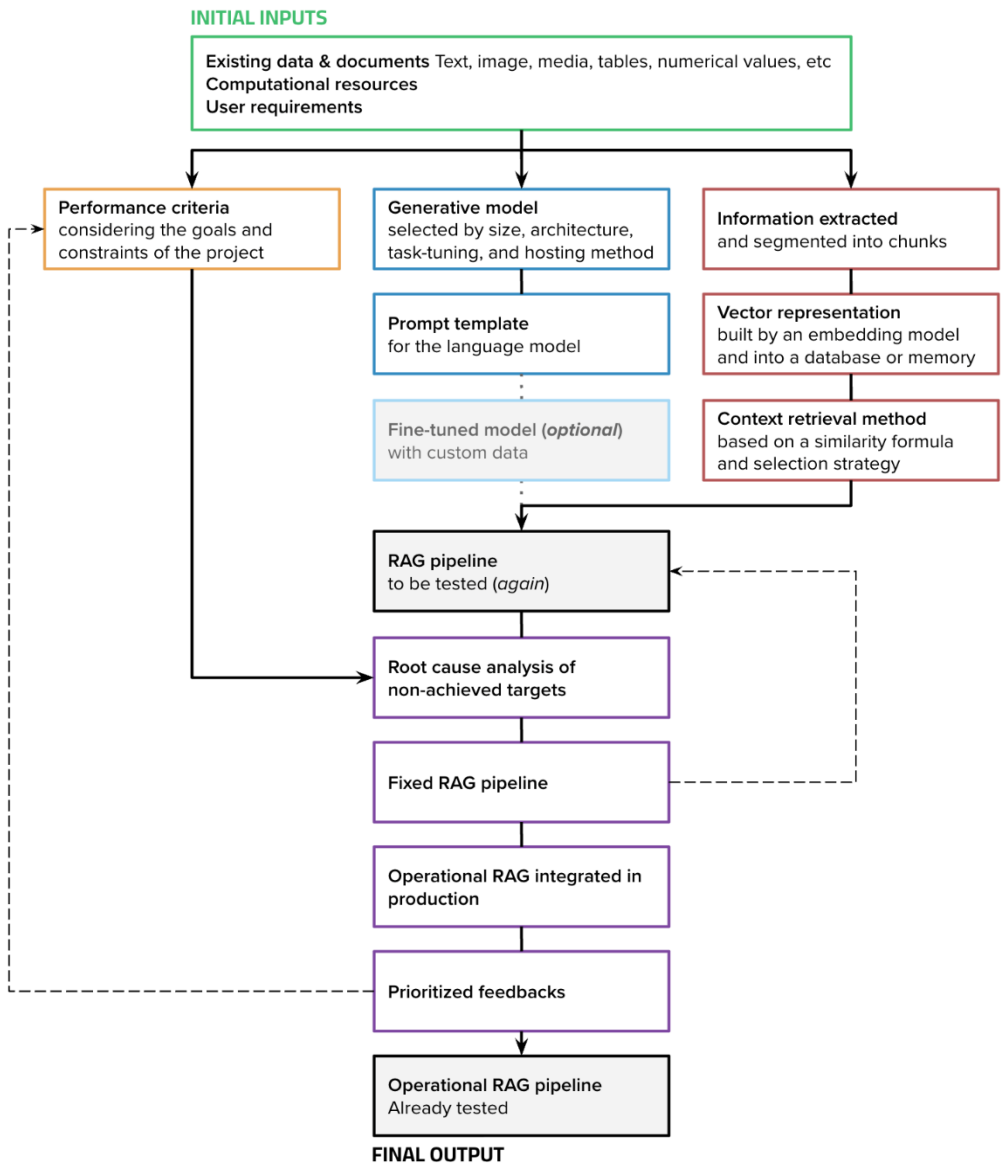

*Figure 2 : Input-output relationship diagram (information model)*

# 4. Industrial test case

To validate the relevance of the presented method, the EASI-RAG method was applied to a real-world use case in an industrial company.

**4.1 Company and data description**

The industrial partner is an environmental analysis laboratory in France employing around 200 people. The case study focuses on the development of a virtual assistant designed to answer employee questions based on the company's operating procedures.

The team involved in the implementation of this virtual assistant consists of six members, distributed across the previously described roles as follows:
- **Users:** Two employees with different age and seniority within the company were chosen to represent a 15-person team;
- **Data Expert:** a senior team member with over 20 years of experience and extensive knowledge of the operating procedures;
- **Data Owner:** the head of the concerned service; and
- **Developers:** two members of the IT team of the company, composed of five people. None of them had prior experience with RAG systems.

The operating procedures used in this project consist of nine documents: seven Word files and two Excel spreadsheets. These documents vary in length and content. Some include images, while others are purely textual. Most documents are structured, but not all of them. Table 7 provides a summary of the characteristics of each document. For confidentiality reasons, documents are not disclosed.

| Document number | Number of tokens | Number of pages | Type of file | Structured | Includes pictures |
|---|---|---|---|---|---|
| 1 | 950 | 5 | Word | Yes | Yes |
| 2 | 16947 | 95 | Word | Yes | Yes |
| 3 | 2389 | 10 | Word | Yes | Yes |
| 4 | 2458 | 2 | Excel | - | No |
| 5 | 2114 | 6 | Word | Yes | Yes |
| 6 | 2017 | 7 | Word | Yes | Yes |
| 7 | 1096 | 8 | Word | Yes | Yes |
| 8 | 945 | 2 | Word | No | No |
| 9 | 796 | 2 | Excel | - | No |
| **Total** | **29712** | **137** | - | - | - |

*Table 7: Summary of the industrial data used in the use case*

**4.2 Initial Design of the RAG process**

Following the EASI-RAG method, the first step is a quick initial setup of a RAG pipeline, that will be later upgraded through several improvement loops. The initial components of our RAG pipeline are the following:
- **Data retrieval and chunking:** Data are retrieved with standard retrieving tools (LangChain library[1]). No pretreatment is applied to data; notably, no preprocessing is applied to data before feeding them in the RAG pipeline. Given that the data used contains both structured and unstructured documents, the initial chunking consists of a simple segmentation into fixed-size chunks of 1,000 tokens, without overlap.
- **Vectorisation:** Used data contain little highly specific vocabulary, therefore, a dense vectorization is more appropriate. The MTEB is then employed to search for an embedding model, meeting the following criteria: a sufficiently small model size to embed all documents within a reasonable time on a company's computer, and training oriented toward retrieval tasks. Once the search is completed, several models are selected and tested: the model with the highest MTEB score and a reasonable embedding time for the company's requirements is the *all-MiniLM-L6-v2*[2] model.
- **Retrieval of relevant chunks:** As the chunks are of 1,000 tokens each, the number of chunks initially retrieved is set to three, in order to limit the amount of context provided and to maintain better control over the retrieved results at this stage.
- **Prompt engineering:** The initial prompt used in this use case is intentionally simple: *"Using the following contextual elements:* [list of retrieved chunks, separated by "---------------"] *respond to the following query:* [insertion of the query]".
- **Language Model selection:** As the data used does not contain any sensitive corporate information, the selected solution for the language model is "gpt-3.5-turbo"[3], accessed through the OpenAI API. The average cost par query obtained with this model is approximately $0.0013 per query (June 2025).

---
[1] https://python.langchain.com/docs/integrations/document_loaders/

### 4.3 Evaluation of the RAG Process

Once the initial RAG pipeline is setup, the six project participants jointly defined a set of validation criteria for assessing the effectiveness of the RAG system:

- The designed RAG pipeline operates on a "standard" computer, namely one equipped with 8 GB of RAM and no GPU.
- The system produces responses almost instantaneously, with an inference time on the order of one second.
- A set of 42 test queries is then defined **by the users,** reflecting realistic questions that could assist them in everyday situations by enabling faster retrieval of information. Eight additional questions, whose answers are not present in the documents, are added to the set, resulting in a total of 50 test queries. For each test query, the data expert provides the expected answer and its location in the documents to facilitate the verification by the developers during subsequent stages.
- For each question, two performance criteria are defined: including all expected elements (good recall) and containing no incorrect information (good precision). A response is considered correct if it satisfies both criteria. Above all, the RAG pipeline must **never** give instructions that contradict operational procedures. Therefore, the following target values are defined:
    - More than 80% correct responses;
    - Fewer than 20% acceptable responses (partial responses or "I don't know" statements); and
    - Zero responses that contradict operational manual instructions.

With these criteria established, the team proceeds to evaluate the RAG pipeline against the defined targets. The initial evaluation of the RAG pipeline produced the following outcomes:
- Average response time: approximately 2 seconds;
- 17 correct answers;
- 19 acceptable answers; and
- 14 incorrect answers.

### 4.5 Error analysis and correction

The incorrect responses are subsequently analysed following the methodology outlined in Section 3. The analysis reveals ten distinct causes of error, which are summarised and illustrated in the following Figure 3.

---

[2] https://huggingface.co/sentence-transformers/all-MiniLM-L6-v2
[3] https://platform.openai.com/docs/models/gpt-3.5-turbo

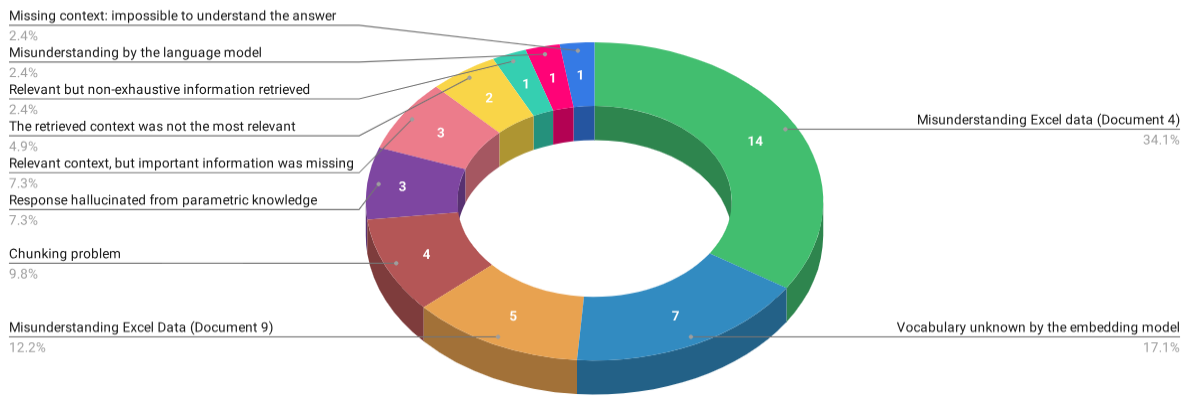

*Figure 3: Pareto analysis of encountered errors after initial deployment*

The cumulative number of error causes shown in the chart exceeds the total number of incorrect results, as multiple issues can contribute to a single incorrect outcome.

Identified errors have been corrected through six successive modifications, addressing the issues from the most frequent to the least frequent:

**Misunderstanding Excel data - Document 4:** Without preprocessing, the chunks obtained from Document 4 were unusable: the first chunk contained column headers and initial rows, while subsequent chunks contained following rows without headers. Additionally, chunks combined unrelated topics, which led to poor semantic understanding of obtained chunks and irrelevant retrievals. To address this, a preprocessing step was added to split this document into one row per chunk and prefix each cell's content with its column header.

**Vocabulary unknown by the embedding model:** Some user queries contained rare keywords that the embedder effectively "ignored" by blending their representation with that of surrounding terms. As a result, retrieved chunks were off-topic, driven by similarity to non-relevant context words. To mitigate this, a hybrid vectorization approach combining dense embeddings with sparse BM25 was implemented. The top three chunks from each method are retrieved for improved relevance.

**Misunderstanding Excel Data (Document 9):** Similar to Document 4, data from Document 9—another Excel file— could not be retrieved. However, the previous solution failed here due to different structures in documents: Document 9 contains merged cells and mostly consists in empty content, with information conveyed by crosses or colours in some cells. Given its brevity, the table was manually converted into a textual description (one phrase for each raw), allowing easier retrieval.

**Chunking problem:** Some chunks were incomplete or split within key sentences, making retrieval more difficult. To address this, documents were segmented according to their internal structure (sections and subsections) when possible, rather than using a fixed 1000-token size. Only one document (Document 8) lacked such structure. As it contained fewer than 1000 tokens, it was kept as a single chunk.

**Response hallucinated from parametric knowledge:** Some responses were not based on the information contained in the retrieved chunks, but instead on the parametric knowledge of the language model. To address this, the prompt was revised to include the following instruction at the end of each query: *"Do not use your prior knowledge; use only the information provided."*

**Relevant context, but important information missing:** In some cases, the RAG pipeline retrieves only chunks that are relevant to the query but still miss other useful ones. This issue stems from insufficient context. To address it, the amount of context provided was increased to retain the top five chunks from each vectorisation method (sparse and dense).

These various modifications resulted in the following outcomes:
- Average response time: approximately 2 seconds (unchanged);
- 44 correct answers;
- 7 acceptable answers; and
- 0 incorrect answers.

It is important to emphasise that the implemented modifications do not necessarily produce uniformly positive effects across all queries. For example, increasing the amount of context can enhance the accuracy of certain responses while inadvertently diminishing that of others. In our case, the inclusion of the instruction *"Do not use your prior knowledge; use only the information provided."* led to an overall improvement in answer quality but also resulted in a decline for three queries: the virtual assistant, which had previously returned the correct answers, now responded with an explicit statement of uncertainty. This highlights the need to assess each change not only in terms of aggregate performance gains, but also with regard to its potential adverse effects on individual cases.

### 4.7 Integration and User Feedback

After implementing the previous corrections, the tool was deployed into production. A minimalist interface was created, and the tool was integrated into a software toolkit accessible to the team members. A *"Report Incorrect Answer"* button was added to streamline feedback: when clicked, the query and its response are sent to the data expert, who (1) provides the correct answer to the user, and (2) verifies whether the dataset contains it. If absent, the dataset is supplemented and updated; if present, the case is forwarded to developers for investigation.

A weekly feedback meeting was also established to review user comments systematically. After two weeks of production use, feedback indicated that the tool generally delivers helpful answers. However, users noted that responses did not display the contextual excerpts used, limiting their ability to verify or explore the information further. They recommended showing the relevant excerpt and source document. Interface improvements were also suggested to enhance user experience. Additionally, two new test questions—both producing incorrect answers—were identified in the first week. These were added to the official test set to ensure that future enhancements address the issues effectively.

All these elements, from the kickoff meeting to production deployment of the first version, were implemented within three weeks.

### 4.8 Overview and contributions of the method

Applying the proposed method to a real-world case enabled the deployment of a RAG system in just a few weeks, using developers with no prior RAG experience. The deployed system achieved 86% correct answers and produced no unsatisfactory responses. Its agile nature saved significant time by avoiding unnecessary complex techniques and focusing on high-priority issues, allowing rapid delivery of a tool that supports the majority of cases. Iterative improvements targeted critical blockers while avoiding effort on non-essential features. Specifically, this approach prevents the need in this case for:

- Complex chunking — a simple hierarchy-based method was sufficient, more advanced chunking techniques (semantic or combined chunking) were unnecessary;
- Multi-modal RAG — despite the presence of figures, testing showed that accessing image-based information was not required since the text contained adequate explanations to answer questions
- Advanced retrieval techniques: while hybrid vectorisation was needed to retrieve chunks linked to specific vocabulary terms, other sophisticated retrieval methods (e.g., Hypothetical Document Embedding, reranking, child-parent retrieving) were not; and
- Enhanced model reasoning — minor prompt adjustments sufficed without advanced reasoning strategies.

Hardware costs were minimal (a few tens of euros for LLM usage). The main cost was human time: two part-time developers over three weeks (approximately 50 hours), plus about 5 hours each from users, the data expert, and the process owner for meetings, totalling around 70 person-hours. This equates to one full-time employee for two weeks — a reasonable investment even for SMEs.

The results show that the method enables efficient RAG deployment in SMEs by focusing resources on essential features, delivering strong performance while keeping both complexity and cost low.

## 5. Discussion

The method proposed in this paper has proven to be both efficient and accessible, enabling the rapid deployment of a RAG system in a real-world setting using a small team with no experience in RAG technologies. Its agile and pragmatic design allows for the quick development of a functional tool that delivers correct answers in over 85% of cases in a real industrial use case, without requiring advanced techniques such as complex chunking, multimodal inputs, or sophisticated language model reasoning. By focusing on the most critical issues and iterating based on user feedback, the method ensures relevant performance while minimizing unnecessary development work. Additionally, the implementation cost—especially in terms of hardware and human effort—remains low, making this approach particularly suitable for SMEs looking to adopt AI solutions without significant resource investment.

The work presented in this paper presents certain limitations. The proposed method has only been tested on a single real-world case, with specific data and particular stakeholders. Its applicability to other contexts with different datasets or teams has not yet been explored, and broader validation across diverse applications and profiles would be beneficial. Moreover, the study focuses solely on RAG techniques, excluding other NLP methods such as fine-tuning language or embedding models, or the full RAG pipeline. While more complex and resource-intensive, these approaches can improve performance by incorporating domain-specific knowledge and adapting to varied document types. They were excluded due to SME constraints but remain valuable alternatives when RAG alone is insufficient.

Implementing RAG systems in industrial settings greatly enhances information access for staff and clients, offering key benefits: saving time by retrieving relevant data within seconds instead of manually searching large document volumes, and improving quality by uncovering answers that might otherwise remain unnoticed. These tools are particularly valuable for enterprises, especially when training new employees who require frequent information access. However, adoption requires careful monitoring of two risks: reduced interpersonal communication, potentially devaluing roles centered on knowledge sharing, and over-reliance on automated assistance, which may erode manual search skills. Maintaining some manual retrieval capabilities is essential to preserve expertise and prevent excessive dependency on the system.

## 6. Conclusion

The EASI-RAG method presented in this paper describes an agile approach focused on the rapid deployment of RAG tools within the context of industrial SMEs. This method outlines the various activities required to implement RAG tools, specifying for each activity the possible techniques, the personnel involved, and the expected deliverables, following the framework of Zellner's methodology. A key aspect of this method is its agile nature, which involves quickly deploying an initial version of the tool and then progressively improving it through iterative feedback loops. The paper also proposes a list of techniques applicable depending on the challenges encountered during the initial implementation of a RAG tool.

The proposed conceptual method was partially validated through a real-world application in an industrial SME. Its use enabled the rapid deployment of a RAG tool to answer operators' questions based on operational procedures. Implementation was achieved at low cost, with a development team with no prior RAG expertise, while maintaining document confidentiality. Therefore, this method provides an ideal resource for SMEs, where flexibility is a major advantage and significant human or material investment may hinder the development of new tools. Furthermore, the continuous involvement of end-users throughout the project facilitated quick adoption by the production team.

Future research could broaden validation of the EASI-RAG method across diverse SMEs to assess generalisability and robustness. Longitudinal studies should examine long-term effects on human factors, including communication and information-retrieval skills. Integrating EASI-RAG with industrial systems (ERP, MES, maintenance platforms) could enhance interoperability, while partially automating tasks such as document ingestion, feedback analysis, and technique selection may reduce deployment time and effort. These developments would make the method more accessible to resource-constrained SMEs and strengthen its practical applicability.

**Acknowledgements**

The authors would like to thank the industrial partner who provided access to the data and the real-world use case that made this work possible. Their support and collaboration were instrumental in grounding our research in practical applications.

**Declaration of Interest**

The authors declare that they have no conflict of interest, and that no financial, commercial, or personal relationships influenced the design, execution, or reporting of this work.